# Mitigating Gender Bias in Natural Language Processing: Literature Review

**Tony Sun**[*][†], **Andrew Gaut**[*][†], **Shirlyn Tang**[†], **Yuxin Huang**[†],
**Mai ElSherief**[†], **Jieyu Zhao**[‡],
**Diba Mirza**[†], **Elizabeth Belding**[†], **Kai-Wei Chang**[‡], and **William Yang Wang**[†]

[†]Department of Computer Science, UC Santa Barbara
[‡]Department of Computer Science, UC Los Angeles
{tonysun, ajg, shirlyntang, yuxinhuang}@ucsb.edu
{mayelsherif, dimirza, ebelding, william}@cs.ucsb.edu
{jyzhao, kwchang}@cs.ucla.edu

## Abstract

As Natural Language Processing (NLP) and Machine Learning (ML) tools rise in popularity, it becomes increasingly vital to recognize the role they play in shaping societal biases and stereotypes. Although NLP models have shown success in modeling various applications, they propagate and may even amplify gender bias found in text corpora. While the study of bias in artificial intelligence is not new, methods to mitigate gender bias in NLP are relatively nascent. In this paper, we review contemporary studies on recognizing and mitigating gender bias in NLP. We discuss gender bias based on four forms of representation bias and analyze methods recognizing gender bias. Furthermore, we discuss the advantages and drawbacks of existing gender debiasing methods. Finally, we discuss future studies for recognizing and mitigating gender bias in NLP.

## 1 Introduction

Gender bias is the preference or prejudice toward one gender over the other (Moss-Racusin et al., 2012). Gender bias is exhibited in multiple parts of a Natural Language Processing (NLP) system, including the training data, resources, pretrained models (e.g. word embeddings), and algorithms themselves (Zhao et al., 2018a; Bolukbasi et al., 2016; Caliskan et al., 2017; Garg et al., 2018). NLP systems containing bias in any of these parts can produce gender biased predictions and sometimes even amplify biases present in the training sets (Zhao et al., 2017).

The propagation of gender bias in NLP algorithms poses the danger of reinforcing damaging stereotypes in downstream applications. This has real-world consequences; for example, concerns have been raised about automatic resume filtering systems giving preference to male applicants when the only distinguishing factor is the applicants' gender.

[*] Equal Contribution.

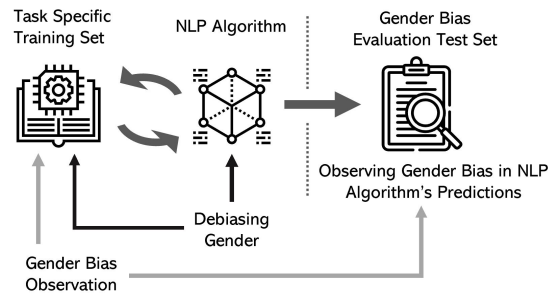

Figure 1: Observation and evaluation of gender bias in NLP. Bias observation occurs in both the training sets and the test sets specifically for evaluating the gender bias of a given algorithm's predictions. Debiasing gender occurs in both the training set and within the algorithm itself.

One way to categorize bias is in terms of allocation and representation bias (Crawford, 2017). Allocation bias can be framed as an economic issue in which a system unfairly allocates resources to certain groups over others, while representation bias occurs when systems detract from the social identity and representation of certain groups (Crawford, 2017). In terms of NLP applications, allocation bias is reflected when models often perform better on data associated with majority gender, and representation bias is reflected when associations between gender with certain concepts are captured in word embedding and model parameters. In Table 1, we categorize common examples of gender bias in NLP following Crawford (2017).

| Task | Example of Representation Bias in the Context of Gender | D | S | R | U |
|---|---|---|---|---|---|
| Machine Translation | Translating "He is a nurse. She is a doctor." to Hungarian and back to English results in "She is a nurse. He is a doctor." (Douglas, 2017) | | ✓ | ✓ | |
| Caption Generation | An image captioning model incorrectly predicts the agent to be male because there is a computer nearby (Burns et al., 2018). | | ✓ | ✓ | |
| Speech Recognition | Automatic speech detection works better with male voices than female voices (Tatman, 2017). | | | ✓ | ✓ |
| Sentiment Analysis | Sentiment Analysis Systems rank sentences containing female noun phrases to be indicative of anger more often than sentences containing male noun phrases (Park et al., 2018). | | ✓ | | |
| Language Model | "He is doctor" has a higher conditional likelihood than "She is doctor" (Lu et al., 2018). | | ✓ | ✓ | ✓ |
| Word Embedding | Analogies such as "man : woman :: computer programmer : homemaker" are automatically generated by models trained on biased word embeddings (Bolukbasi et al., 2016). | ✓ | ✓ | ✓ | ✓ |

Table 1: Following the talk by Crawford (2017), we categorize representation bias in NLP tasks into the following four categories: (D)enigration, (S)tereotyping, (R)ecognition, (U)nder-representation.

Briefly, denigration refers to the use of culturally or historically derogatory terms; stereotyping reinforces existing societal stereotypes; recognition bias involves a given algorithm's inaccuracy in recognition tasks; and under-representation bias is the disproportionately low representation of a specific group. We identify that both allocative and representational harms often arise in NLP systems due to statistical patterns in the training corpora, which are then embedded in semantic representations and the model.

Gender bias in NLP is a complex and compound issue, requiring interdisciplinary communication. As NLP systems have been increasingly integrated with our daily life thanks to modern AI developments, we need both immediate solutions to patch current systems as well as fundamental approaches to debias. In this paper, we provide a comprehensive literature review to summarize recent attempts for recognizing and mitigating bias in NLP systems. Most debiasing methods can be depicted as a special case in Figure 1.

We make two primary contributions. (1) We summarize recent studies of algorithmic bias in NLP under a unified framework for the ease of future discussion. (2) We critically discuss issues with current debiasing methods with the purpose of identifying optimizations, knowledge gaps, and directions for future research.

## 2 Observing Gender Bias

Recent work in analyzing gender bias in NLP has focused on quantifying bias through psychological tests, performance differences between genders for various tasks, and the geometry of vector spaces. We provide an overview of gender bias evaluation methods and discuss types of representation bias each method identifies.

### 2.1 Adopting Psychological Tests

In psychology, the Implicit Association Test (IAT) is used to measure subconscious gender bias in humans, which can be quantified as the difference in time and accuracy for humans to categorize words as relating to two concepts they find similar versus two concepts they find different (Greenwald et al., 1998; Caliskan et al., 2017). For instance, to measure subconscious associations of genders with arts and sciences, participants are asked to categorize words as pertaining to (males or the sciences) or (females or the arts) (Nosek et al., 2009). The participants are then asked to categorize words as pertaining to (males or the arts) or (females or the sciences). If participants answered faster and more accurately in the former setting, it indicates that humans subconsciously associate males with the sciences and females with the arts.

Caliskan et al. (2017) adopt the IAT's core concept, measuring gender bias through the difference in strength of association of concepts, to measure bias in word embeddings using the Word Embedding Association Test (WEAT) (Caliskan et al., 2017). The authors confirm that human biases found through IAT tests exist in GloVe and Word2Vec embeddings. Finally, the authors demonstrate a positive correlation between the strength of association of an occupation word embedding with the female gender and the percentage of females in that occupation in United States, with the percentages taken from Bureau of Labor Statistics labor force participation data. Notably, Garg et al. (2018) show that bias in word embeddings can be used to track social changes such as increased or decreased female participation in the workforce. May et al. (2019) extend WEAT to create the Sentence Encoder Association Test (SEAT), capable of testing sentence encoders (e.g., ELMo (Peters et al., 2018)) for human biases found in IAT tests.

## 2.2 Analyzing Gender Sub-space in Embeddings

Bolukbasi et al. (2016) define gender bias as the correlation between the magnitude of the projection onto the gender subspace of a word embedding representing a gender-neutral word and that word's bias rating, as rated by crowd workers. To identify the gender subspace, they first build a linear support vector machine to classify words into a set of gender-specific and a set of gender-neutral words based on a training set of hand-selected gender-specific words. The authors then identify a gender direction by aggregating ten gender pairs (e.g. she-he, her-his, woman-man, etc.) and using principal component analysis to find a single eigenvector that exhibits significantly greater variance than the rest. Manzini et al. (2019) extend this method and their approach can be used to find non-binary gender bias by aggregating n-tuples instead of gender pairs.

However, Gonen and Goldberg (2019) note that the above method fails to capture the full picture of gender bias in vector spaces. Specifically, even after the projections of word embeddings representing gender-neutral words onto the gender subspace have been removed, word embeddings representing words with similar biases still cluster together. They further introduce the notion of cluster bias. Cluster bias of a word $w$ can be measured as the percentage of male or female stereotypical words among the $k$ nearest neighbors of $w$'s embedding where the male or female stereotypical words are obtained through human annotation.

## 2.3 Measuring Performance Differences Across Genders

In most NLP tasks, a model's prediction should not be heavily influenced by the gender of the entity mentions or contexts in the input. To evaluate whether or not this is the case, consider two sentences that act as the inputs to a model for which the only differences are the words that correspond to gender, such as "*He* went to the park" vs "*She* went to the park". We refer to changing the gender of the gendered nouns as *gender-swapping*. Gender-swapping can be generalized to sentences by swapping each male-definitional word with its respective female equivalent and vice-versa (Zhao et al., 2018a; Lu et al., 2018; Kiritchenko and Mohammad, 2018). If the model does not make decisions based on genders, it should perform equally for both sentences. Otherwise, the difference in evaluation scores reflects the extent of gender bias found in the system.

For example, Dixon et al. (2017) introduce two metrics to measure these performance differences – False Positive Equality Difference (FPED) and False Negative Equality Difference (FNED) – that have been used to measure gender bias in abusive language detection (Park et al., 2018). These are defined as the differences in the false positive and false negative rates, respectively, of predictions of a model between original and gender-swapped inputs. We note that these measurements can generalize to tasks aside from abusive language detection.

By designing test sets, measuring performance differences between genders reveals representational gender bias in the context of recognition, stereotyping, and under-representation. If, for instance, an image captioning model is worse at recognizing a *woman* than a *man* when they are each sitting in front of a computer (Burns et al., 2018), that is a clear indicator of recognition bias. If this prediction inaccuracy arises as a consequence of the algorithm's association between *man* and *computer*, then this example also reveals stereotyping in the image captioning model. One can also imagine that if the model is not debiased and these errors propagate over a large sample of images, then the model may further contribute to the under-representation of minority.

Standard evaluation data sets in NLP are inadequate for measuring gender bias. For one, these data sets often also contain biases (such as containing more male references than female references), so evaluation on them might not reveal gender bias. Furthermore, predictions made by systems performing complex NLP tasks depend on many factors; we must carefully design data sets to isolate the effect of gender of the output in order to be able to probe gender bias. We name these data sets Gender Bias Evaluation Testsets (GBETs).

The goal of designing GBETs is to provide check that NLP systems avoid making mistakes due to gender bias. Some may argue that the artificial design of GBETs does not reflect the true distribution of the data, implying that these evaluations are artificial. We argue that if humans can avoid making mistakes due to gender bias, then machines should as well. Additionally, systems that make biased predictions may discourage minorities from using those systems and having their data collected, thus worsening the disparity in the data sets (Hashimoto et al., 2018). We provide an overview of publicly available GBETs in Table 2.

| Data Set | Task | Probing Concept | Size |
| --- | --- | --- | --- |
| Winogender Schemas (Rudinger et al., 2018) | Coreference Resolution | Occupation | 720 English Sentences |
| WinoBias (Zhao et al., 2018a) | Coreference Resolution | Occupation | 3,160 English Sentences |
| GAP (Webster et al., 2018) | Coreference Resolution | Names | 4,454 English Contexts |
| EEC (Kiritchenko and Mohammad, 2018) | Sentiment Analysis | Emotion | 8,640 English Sentences |

Table 2: Summary of GBETs. GBETs evaluate models trained for specific tasks for gender bias. GBETs use differences in values of the probing concept or prediction accuracies relating to the probing concept between gender-swapped data points to measure bias.

**Gender-swapped GBETs:** In the following, we review GBETs in coreference resolution and sentiment analysis applications.

For coreference resolution, Rudinger et al. (2018) and Zhao et al. (2018b) independently designed GBETs based on Winograd Schemas. The corpus consists of sentences which contain a gender-neutral occupation (e.g., doctor), a secondary participant (e.g., patient), and a gendered pronoun that refers either the occupation or the participant. The coreference resolution system requires the identification of the antecedent of the pronoun. For each sentence, Rudinger et al. (2018) consider three types of pronouns (female, male, or neutral), and Zhao et al. (2018b) consider male and female pronouns. The two datasets have a few notable differences (see the discussion in (Rudinger et al., 2018)).

Note that simply measuring a global difference in accuracies of a model between inputs with different gendered pronouns is insufficient. For example, a model could predict females and males to be coreferent to "secretary" with 60% and 20% accuracy, respectively. If that same model predicts females and males coreferent to "doctor" with 20% and 60% accuracy, respectively, then the global average accuracy for each gender is equivalent, yet the model exhibits bias.[1] Therefore, Zhao et al. (2018b) and Rudinger et al. (2018) design metrics to analyze gender bias by examining how the performance difference between genders with respect to each occupation correlate with the occupational gender statistics from the U.S Bureau of Labor Statistics.

Another GBET for coreference resolution named GAP contains sentences mined from Wikipedia and thus can perform an evaluation with sentences taken from real contexts as opposed to artificially generated ones (Webster et al., 2018). GAP does not include stereotypical nouns; instead, pronouns refer to names only. Gender bias can be measured as the ratio of $F_1$ scores on inputs for which the pronoun is female to inputs for which the pronoun is male. Notably, sentences are not gender-swapped, so there may be differences in difficulty between sentences in male and female test sets.

For sentiment analysis, a GBET dataset named *Equity Evaluation Corpus* (EEC) (Kiritchenko and Mohammad, 2018) is designed. Each EEC sentence contains an emotional word (e.g., anger, fear, joy, sadness), with one of five intensities for each emotion and a gender-specific word. Gender bias is measured as the difference in emotional intensity predictions between gender-swapped sentences.

## 3 Debiasing Methods Using Data Manipulation

Several approaches have been proposed for debiasing gender stereotypes in NLP by working on two tangents: (1) text corpora and their representations and (2) prediction algorithms. In this section, we will discuss the techniques to debias text corpora and word embeddings. We do the same for techniques to mitigate gender bias in algorithms in Section 4.

We note that debiasing methods can be categorized as retraining and inference (see Table 3). Retraining methods require that the model is trained again, while inference methods reduce bias without requiring the existence of the original training set. Retraining methods tend to address gender bias in its early stages or even at its source. However, retraining a model on a new data set can be costly in terms of resources and time. Inference methods, on the other hand, do not require models to be retrained; instead, they patch existing models to adjust their outputs providing a testing-time debiasing. We will discuss different debiasing methods from these two perspectives.

---

[1] For the sake of simplicity, we illustrate the motivation in accuracy. The coreference resolution systems may be evaluated using a different metric.

## 3.1 Debiasing Training Corpora

We review three approaches for debiasing gender in the literature.

### 3.1.1 Data Augmentation

Oftentimes a data set has a disproportionate number of references to one gender (e.g. OntoNotes 5.0) (Zhao et al., 2018a). To mitigate this, Zhao et al. (2018a) proposed to create an augmented data set identical to the original data set but biased towards the opposite gender and to train on the union of the original and data-swapped sets. The augmented data set is created using gender-swapping. This is similar to the method used to create GBETs; however, the goal of data augmentation is to debias predictions by training the model on a gender-balanced data set, while GBETs are created specifically to evaluate the gender bias of those predictions both before and after debiasing.

Data augmentation works as follows: for every sentence in the original data set, create that sentence's gender-swapped equivalent using the procedure described in 2.3. Next, apply name-anonymization to every original sentence and its gender-swapped equivalent. Name anonymization consists of replacing all named entities with anonymized entities, such as "E1". For instance, *Mary likes her mother Jan* becomes *E1 likes his father E2* after applying gender-swapping and name anonymization for data augmentation. This removes gender associations with named entities in sentences. The model is then trained on the union of the original data set with name-anonymization and the augmented data set. The identification of gender-specific words and their equivalent opposite gender word requires lists typically created by crowd workers.

Data augmentation has been shown to be flexible; it can mitigate gender bias in several different models in many different tasks. When applied to a neural network based coreference resolution model (Lee et al., 2017, 2018) originally trained on OntoNotes 5.0 which was tested on WinoBias, gender augmentation lowered the difference between $F_1$ scores on pro-stereotypical and anti-stereotypical test sets significantly, which indicates the model was less inclined to make gender-biased predictions (Zhao et al., 2018a, 2019). In hate speech detection, data augmentation reduced FNED and FPED differences between male and female predictions of a Convolutional Neural Network by a wide margin (Park et al., 2018). Data augmentation without name-anonymization has also been used to debias knowledge graphs built from Bollywood movie scripts (Madaan et al., 2018) by swapping the nodes for the lead actor and actress, but metrics evaluating the success of gender-swapping were not provided.

| Methods | Method Type |
|---|---|
| Data Augmentation by Gender-Swapping | Retraining |
| Gender Tagging | Retraining |
| Bias Fine-Tuning | Retraining |
| Hard Debiasing | Inference |
| Learning Gender-Neutral Embeddings | Retraining |
| Constraining Predictions | Inference |
| Adjusting Adversarial Discriminator | Retraining |

Table 3: Debiasing methods can be categorized according to how they affect the model. Some debiasing methods require the model to be retrained after debiasing (Retraining). Others modify existing models' predictions or representations (Inference).

Data augmentation is easy to implement, but creating the annotated list can be expensive if there is high variability in the data or if the data set is large since more annotations will be required. Furthermore, data augmentation doubles the size of the training set, which can increase training time by a factor specific to the task at hand. Lastly, blindly gender-swapping can create nonsensical sentences – for example, gender-swapping "*she* gave birth" to "*he* gave birth" (Madaan et al., 2018).

### 3.1.2 Gender Tagging

In some tasks, like Machine Translation (MT), confounding the gender of the source of a data point can lead to inaccurate predictions. Current MT models predict the source to be male a disproportionate amount of time (Prates et al., 2018; Vanmassenhove et al., 2018). This happens because training sets are dominated by male-sourced data points, so the models learn skewed statistical relationships and are thus more likely to predict the speaker to be male when the gender of the source is ambiguous (Vanmassenhove et al., 2018).

Gender tagging mitigates this by adding a tag indicating the gender of the source of the data point to the beginning of every data point. For instance, "I'm happy" would change to "MALE I'm happy." In theory, encoding gender information in sentences could improve translations in which the gender of the speaker affects the translation (i.e. "I am happy" could translate to "Je suis heureux" [M] or "Je suis heureuse" [F]), since English does not mark the gender of the speaker in this case. The tag is then parsed separately from the rest of the data by the model. The goal is to preserve the gender of the source so the model can create more accurate translations (Vanmassenhove et al., 2018).

Gender tagging is effective: a Sequence-to-Sequence Neural Network trained on Europarl increased BLEU scores significantly for machine translations from English to French in which the first-person speaker was female (Vanmassenhove et al., 2018). Sentences with male first-person speakers had accuracy increases by a sizeable margin. However, gender-tagging can be expensive: knowing the gender of the source of a data point requires meta-information, and obtaining this could be costly in terms of memory usage and time. Furthermore, MT models may need to be redesigned to correctly parse the gender tags.

### 3.1.3 Bias Fine-Tuning

Unbiased data sets for a given task may be scarce, but there may exist unbiased data sets for a related task. Bias fine-tuning incorporates transfer learning from an unbiased data set to ensure that a model contains minimal bias before fine-tuning the model on a more biased data set used to train for the target task directly (Park et al., 2018). This allows models to avoid learning biases from training sets while still being adequately trained to perform a task.

Bias fine-tuning has been shown to be relatively effective. Park et al. (2018) use transfer learning from a gender unbiased abusive tweets data set (Founta et al., 2018) and fine-tuning on a gender-biased sexist tweets data set (Waseem and Hovy, 2016) to train a Convolutional Neural Network (CNN). They evaluate the CNN using a GBET evaluation set (which is private, so not mentioned in 2.3). They tested the same model after training it on gender-swapped data sets as well. Park et al. (2018) find that gender-swapping was more effective at both removing bias and retaining performance than bias fine-tuning. However, transfer learning may have been ineffective in this case because abusive language detection data sets and sexist language detection data sets have significant differences. For more similar data sets, bias fine-tuning may be more effective; further testing is necessary.

## 3.2 Debiasing Gender in Word Embeddings

Word embeddings represent words in a vector space. These embeddings have been demonstrated to reflect societal biases and changing views during social movements in the United States (Garg et al., 2018).

As the word embedding model is a fundamental component in many NLP systems, mitigating bias in embeddings plays a key role in the reduction of bias that is propagated to downstream tasks (e.g., (Zhao et al., 2018a)). However, it is debatable if debiasing word embeddings is a philosophically right step towards mitigating bias in NLP. Caliskan et al. (2017) argue that debiasing word embeddings blinds an AI agent's perception rather than teaching it to perform fair actions. We refer readers to the discussion in (Caliskan et al., 2017).

It is also important to recognize that removing gender bias from the embedding space entirely is difficult. While existing methods successfully mitigate bias with respect to projection onto the gender subspace in some degrees, Gonen and Goldberg (2019) show that gender bias based on more subtle metrics such as cluster bias still exist.

In the following we review two families of approaches to debias gender in word embeddings. One difference between these two types of methods is that the former does not require retraining embeddings, whereas the latter does.

### 3.2.1 Removing Gender Subspace in Word Embeddings

Schmidt (2015) first removed similarity to the gender subspace in word embeddings by building a genderless framework using cosine similarity and orthogonal vectors (Schmidt, 2015). Removing the gender component, though, pushes the word *he* to become the 6th closest word to *she* when it was the 1,826th closest previously. The genderless framework may be flawed because the semantic definition of a given word may be closely tied to its gender component. However, a case can also be made that a word's gender component should play a key role in its semantic definition. We encourage future work to collaborate with social scientists for further discussion on this topic.

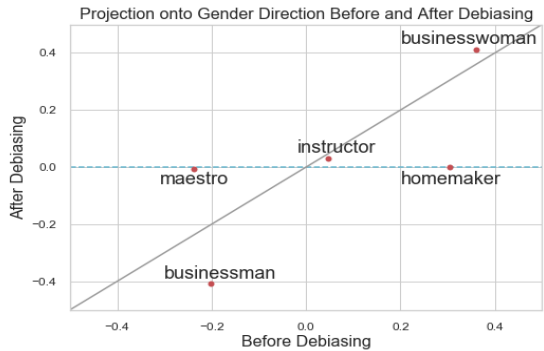

Figure 2: We project five word2vec embeddings onto the 'he' - 'she' direction before and after neutralizing the gender-neutral words *maestro, instructor*, and *homemaker* and equalizing the gender-specific pair *businessman* and *businesswoman* (Bolukbasi et al., 2018). For both x and y-axes, negative values represent male gender bias and positive values represent female gender bias.

Bolukbasi et al. (2016) build upon Schmidt (2015) and propose to surgically alter the embedding space by removing the gender component only from gender-neutral words. Instead of removing gender altogether, debiasing involves making gender-neutral words orthogonal to the gender direction (see Figure 2). Ultimately, word embeddings with reduced bias performed just as well as unaltered embeddings on coherence and analogy-solving tasks (Bolukbasi et al., 2016).

### 3.2.2 Learning Gender-Neutral Word Embeddings

Zhao et al. (2018b) propose a new method called GN-GloVe that does not use a classifier to create a set of gender-specific words. The authors train the word embeddings by isolating gender information in specific dimensions and maintaining gender-neutral information in the other dimensions. They do this by (1) minimizing the negative difference (i.e. maximizing the difference) between the gender dimension in male and female definitional word embeddings and (2) maximizing the difference between the gender direction and the other neutral dimensions in the word embeddings. This allows for greater flexibility; the gender dimensions can be used or neglected.

Finally, we note that both aforementioned approaches (Bolukbasi et al., 2016; Zhao et al., 2018b) used to debias word embeddings may not work with embeddings in a non-Euclidean space, such as Poincare embeddings (Nickel and Kiela, 2017), because the notion of cosine similarity would no longer apply. Also, it is unclear if these approaches can be extended to other languages beyond English, especially for languages with grammatical genders.

## 4 Debiasing by Adjusting Algorithms

Some gender debiasing methods in NLP adjust predictions in NLP systems. We call these algorithm adjustment methods. In this section, we discuss two such approaches.

### 4.1 Constraining Predictions

Zhao et al. (2017) show that an NLP model risks amplifying bias by making predictions which exacerbate biases present in the training set. For instance, if 80% of coreferents of "secretary" are female in a training set and a model trained on that set predicts 90% of coreferents of "secretary" in a test set to be female, then that model amplifies bias.

Zhao et al. (2017) proposed Reducing Bias Amplification (RBA) based on a constrained conditional model (Roth and Yih, 2004), which takes an existing model's optimization function and constrains that function to ensure its predictions fit defined conditions. For example, when RBA was applied to the visual semantic role labelling (Yatskar et al., 2016), it restricted the ratio of males to females predicted to be doing particular activities to prevent the model from amplifying bias through predictions. The approximate inference can be efficiently solved by Lagrangian relaxation (Rush and Collins, 2012).

### 4.2 Adversarial Learning: Adjusting the Discriminator

Zhang et al. (2018) propose a variation on the traditional generative adversarial network (Goodfellow et al., 2014) by having the generator learn with respect to a protected gender attribute. In other words, the generator attempts to prevent the discriminator from identifying the gender in a given task such as analogy completion. This method has the potential to be generalizable: it can be used to debias any model that uses gradient-based learning.

## 5 Conclusion and Future Directions

In this paper, we summarize recent literature about recognizing and mitigating gender bias in NLP. We acknowledge that the scope of this paper is limited. There is a long history of gender stereotype study in law, psychology, media study, and many other disciplines which we do not discuss. Similar issues of algorithmic bias have also been discussed extensively in artificial intelligence, machine learning, data mining, and several other application fields (e.g., (Calders and Verwer, 2010; Feldman et al., 2015; Hardt et al., 2016; Misra et al., 2016; Kleinberg et al., 2016; Pleiss et al., 2017; Beutel et al., 2017; Misra et al., 2016)). Other important aspects such as model/data transparency (Mitchell et al., 2019; Bender and Friedman, 2018) and privacy preservation (Reddy and Knight, 2016; Elazar and Goldberg, 2018; Li et al., 2018) are also not covered in this literature survey. Besides, we refer the readers to Hovy and Spruit (2016) for a more general discussion of ethical concern in NLP.

The study of gender bias in NLP is still relatively nascent and consequently lacks unified metrics and benchmarks for evaluation. We urge researchers in related fields to work together to create standardized metrics that rigorously measure the gender bias in NLP applications. However, we recognize that different applications may require different metrics and there are trade-offs between different notions of biases (Barocas et al., 2018; Chouldechova and Roth, 2018).

Gender debiasing methods in NLP are not sufficient to debias models end-to-end for many applications. We note the following limitations of current approaches. First, the majority of debiasing techniques focus on a single, modular process of an end-to-end NLP system. It remains to be discovered how these individual parts harmonize together to form an ideally unbiased system. Second, most gender debiasing methods have only been empirically verified in limited applications (Zhang et al., 2018; Zhao et al., 2017), and it is not clear that these methods can generalize to other tasks or models. Third, we note that certain debiasing techniques may introduce noise into a NLP model, causing performance degradation. Finally, hand-craft debiasing approaches may unintentionally encode the implicit bias of the developers.

Below, we identify a few future directions.

**Mitigating Gender Bias in Languages Beyond English.** With few exceptions (Vanmassenhove et al., 2018; Prates et al., 2018), prior work has focused on mitigating gender bias in the English language. Future work can look to apply existing methods or devise new techniques towards mitigating gender bias in other languages as well. However, such a task is not trivial. Methods such as gender-swapping are relatively easy in English because English does not distinguish gender linguistically. However, in languages such as Spanish, each noun has its own gender and corresponding modifiers of the noun need to align with the gender of the noun. To perform gender-swapping in such languages, besides swapping those gendered nouns, we also need to change the modifiers.

**Non-Binary Gender Bias.** With few exceptions (Manzini et al., 2019), work on debiasing in NLP has assumed that the protected attribute being discriminated against is binary. Non-binary genders (Richards et al., 2016) as well as racial biases have largely been ignored in NLP and should be considered in future work.

**Interdisciplinary Collaboration.** As mentioned in Section 1, gender bias is not a problem that is unique to NLP; other fields in computer science such as data mining, machine learning, and security also study gender bias (Calders and Verwer, 2010; Feldman et al., 2015; Hardt et al., 2016; Misra et al., 2016; Kleinberg et al., 2016; Pleiss et al., 2017; Beutel et al., 2017; Kilbertus et al., 2017). Many of these technical methods could be applicable to NLP yet to our knowledge have not been studied.

Additionally, mitigating gender bias in NLP is both a sociological and an engineering problem. To completely debias effectively, it is important to understand how machine learning methods encode biases and how humans perceive biases. A few interdisciplinary studies (Herbelot et al., 2012; Avin et al., 2015; Fu et al., 2016; Schluter, 2018) have emerged, and we urge more interdisciplinary discussions in terms of gender bias. Approaches from other technical fields may improve current debiasing methods in NLP or inspire the development of new, more effective methods even if the properties of the data or problem are different across fields. Discussions between computer scientists and sociologists may improve understanding of latent gender bias found in machine learning data sets and model predictions.

## 6 Acknowledgements

We thank anonymous reviewers for their helpful feedback. We also acknowledge the thoughtful talks in related topics by Kate Crawford, Margaret Mitchell, Joanna J. Bryson, and several others. This material is based upon work supported in part by the National Science Foundation under Grants 1821415 and 1760523.